\documentclass[10pt,twocolumn,letterpaper]{article}

\usepackage{wacv}
\usepackage{color}
\usepackage{times}
\usepackage{epsfig}
\usepackage{graphicx}
\usepackage{amsmath}
\usepackage{amssymb}
\usepackage{algorithm}
\usepackage{algorithmic}
\usepackage{multirow}
\usepackage{subfigure}

\wacvfinalcopy 

\def\wacvPaperID{212} 

\ifwacvfinal\fi
\begin{document}

\title{Efficient Super Resolution Using Binarized Neural Network}

\author{Yinglan Ma\thanks{These two authors contributed equally.} \\
Adobe Inc. Research\\
{\tt\small yingma@adobe.com}
\and
Hongyu Xiong\footnotemark[1] \\
Stanford University\\
{\tt\small hxiong2@stanford.edu}
\and
Zhe Hu \\
Hikvision Research\\
{\tt\small zhe.hu@hikvision.com}
\and
Lizhuang Ma \\
East China Normal University\\
{\tt\small lzma@sei.ecnu.edu.cn}
}

\maketitle
\ifwacvfinal\thispagestyle{empty}\fi

\begin{abstract}
Deep convolutional neural networks (DCNNs) have recently demonstrated high-quality results in single-image super-resolution (SR). 
DCNNs often suffer from over-parametrization and large amounts of redundancy, which results in inefficient inference and high memory usage, preventing massive applications on mobile devices.
As a way to significantly reduce model size and computation time, binarized neural network has only been shown to excel on semantic-level tasks such as image classification and recognition.
However, little effort of network quantization has been spent on image enhancement tasks like SR, as network quantization is usually assumed to sacrifice pixel-level accuracy.
In this work, we explore an network-binarization approach for SR tasks without sacrificing much reconstruction accuracy.
To achieve this, we binarize the convolutional filters in only residual blocks, 
and adopt a learnable weight for each binary filter.
We evaluate this idea on several state-of-the-art DCNN-based architectures, and show that binarized SR networks achieve comparable qualitative and quantitative results as their real-weight counterparts. 
Moreover, the proposed binarized strategy could help reduce model size by $80\%$ when applying on SRResNet~\cite{sisrgan}, and could potentially speed up inference by $5\times$.
\end{abstract}

\section{Introduction}

The highly challenging task of estimating a high-resolution (HR) image from its low-resolution (LR) counterpart is referred to as super-resolution (SR). 
SR, particularly single image SR, has received substantial attention within the computer vision research community, 
and has been widely used in applications ranging from HDTV, surveillance imaging to medical imaging. 
The difficulty of SR is to reconstruct high-frequency details from an input image with only low-frequency information. 
In other words, it is to revert the non-revertible process of low-pass filter and downsampling that produces LR images.
%

Recent literatures have witnessed promising progress of SR using convolutional neural networks~\cite{Dong-PAMI2016,Kim-CVPR2016}. 
However, inefficiency and high model size are big issues in deep neural networks, because of over-parametrization. 
To address these issues in deep neural networks, recently neural network quantization, e.g., using binary weights and operations~\cite{Courbariaux,xnor}, is proposed for semantic-level tasks like classification and recognition.
Binarized values have huge advantages from an efficiency perspective, since multiplications can be eliminated, and bit-wise operations can be used to further reduce computational cost.
Nevertheless, little effort of neural network quantization has been spent on image enhancement tasks like SR, as it was assumed to sacrifice the desired pixel-level accuracy for those tasks.

In this work, we explore a network binarization approach for SR methods.
To our best knowledge it is the first work to explore neural network binarization for image SR task.
We show that simply binarizing the whole SR network does not generate satisfactory results.
Therefore, we propose a binarization strategy for SR task, by (1) applying binarization only to residual blocks, and (2) using learnable weights to binarize convolutional filters.

We apply this strategy to a few state-of-the-art SR neural networks to verify its effectiveness.
The experimental results show that the binarized SR network perform similarly to the real-weight network without sacrificing much image quality, but with significant model size and computational load saving, making it possible to apply state-of-the-art SR algorithms in mobile device applications, video streaming, and Internet-of-Things (IoT) edge-device processing. 

\begin{figure}[ht]
\begin{center}
\begin{tabular}{ccc}
\includegraphics[height=0.31\linewidth]{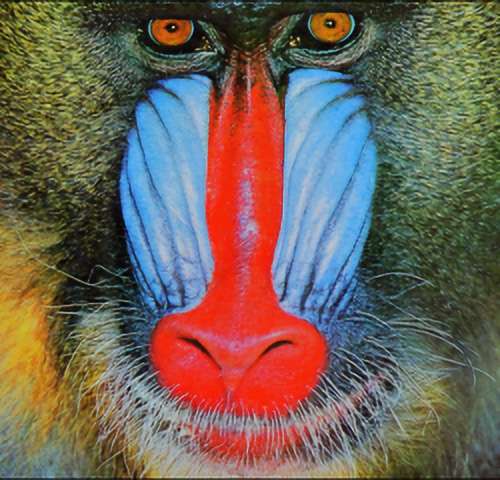} & \hspace{-3mm}
\includegraphics[height=0.31\linewidth]{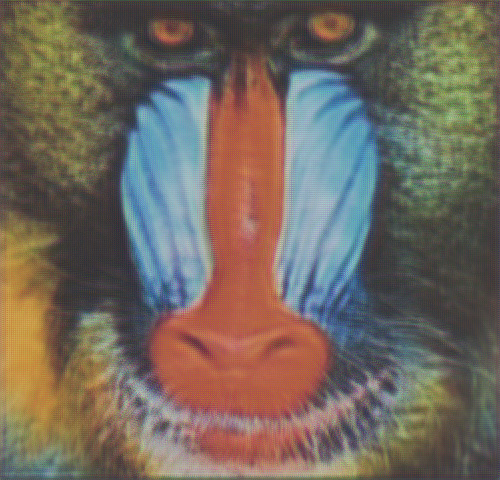} & \hspace{-3mm}
\includegraphics[height=0.31\linewidth]{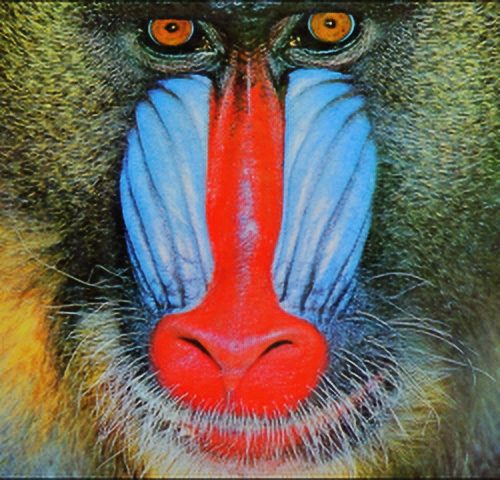} \\
(a) & (b) & (c)\\
\end{tabular}
\caption{Example of binarized networks based on SRResNet~\cite{sisrgan}, with a upscaling factor of $4$. (a) Real-weight network; (b) fully-binarized network directly using binarization strategy~\cite{xnor}; (c) our binarized network.}
\label{ful_binary}
\end{center}
\end{figure}

\section{Related Work}
The recent progresses of single-image super-resolution methods have been well-covered in recent review literatures~\cite{Nasrollahi2014,Yang14_ECCV_Benchmark}. Here we focus our discussion on the most updated SR paradigms based on convolutional neural network (CNN), and progresses in binarized deep neural networks.

{\bf CNN based super resolution.} 
As one of the first proposed SR method based on convolutional structure, SRCNN~\cite{Dong-PAMI2016} trained an end-to-end network using the bicubic upsampling of the LR image as the input. 
The VDSR network~\cite{Kim-CVPR2016} demonstrates significant improvement over SRCNN by increasing the network depth. To facilitate training a deeper model with a fast convergence speed, VDSR aims on predicting the residuals rather than the actual pixel values.
In~\cite{kim2016deeply}, the authors provide a deep convolutional structure that allows a recursive forward propagation of a certain layer, giving decent performance and reducing the number of parameters. 
ESPCN~\cite{Shi2016RealTimeSI} manages to improve the network's performance in both accuracy and speed by directly learning the upsampling filter, and using the LR images as input instead of their bicubic upsampled images. 
The work~\cite{fsrcnn} adopts a similar idea as ESPCN but with more layers and fewer parameters.
DCSCN~\cite{dcscn} proposes a shallower CNN than VDSR, by introducing skip-connections at different levels; directly using LR image as the input, DCSCN investigates and divides the network's function to feature extraction module and reconstruction module, giving one of the highest super-resolution performance.
Johnson et al.~\cite{JohnsonAL16} investigates a perceptual loss related to human perception of high-resolution image, and incorporates the loss with the difference between high-level features from a pretrained VGG network of the predicted and target images. 
SRGAN~\cite{sisrgan} introduces generative adversarial network (GAN)~\cite{gan,dcgan} into SR technique, with a combined content loss and adversarial loss for the perceptual loss function; its generator model, also described as SRResNet, adopts a deep residual framework for feature extraction and uses subpixel-convolutional layer for upscaling reconstruction. 
EDSR ~\cite{lim2017enhanced} moves beyond SRResNet by simplifying the the residual blocks and proposes a multi-scale SR systems, achieving even higher performances. 
LapSRN~\cite{Lai} adopts a Laplacian pyramid to predict HR image; given a fixed upscaling factor for a single level, multiple levels of pyramid could be stacked for larger upscaling factor, and the convolutional filters are shared between different pyramid levels, significantly reducing the number of parameters.

{\bf Neural network with low-precision weights.} 
A great amount of efforts have been made to the speed-up and compression on CNNs during training, feed-forward inference or both stages. Among existing methods, the attempt to restrict CNNs weights into low-precision versions (like binary value or bit-quantized value) attracts great attention from researchers and developers. 
Soudry et al.~\cite{EBP} propose expectation back-propagation (EBP) to estimate the posterior distribution of the weights of the network, which are constrained to $+1$ and $-1$ during feed-forward inference in a probabilistic way. BinaryConnect~\cite{BinaryConnect} extends the idea of EBP, to binarize network weights during training phase directly and updating the real-value weights during the backward pass based on the gradients of the binarized weights. 
BinaryConnect achieves state-of-the-art classification performance for small datasets such as MNIST~\cite{mnist} and CIFAR-10~\cite{krizhevsky2009learning}, showing the possibility that binarized CNNs can have a performance extremely close to real-value network. BinaryNet~\cite{Courbariaux} moves beyond BinaryConnect, whose weights and activations are both binarized. 
XNOR-net~\cite{xnor} extends further beyond BinaryNet and BinaryConnect, by incorporating binarized convolution operation and binarized input during feed-forward inference, showing a significant reduction of memory usage and huge boost of computation speed, despite at certain level compromise of the accuracy. 
Later on, other than binarized networks, a series of efforts have been invested to train CNNs with low-precision weights, low-precision activations and even low-precision gradients, including but not limited to ternary weight network (TWN)~\cite{twn}, DoReFa-Net~\cite{dorefa}, quantized neural network (QNN)~\cite{qnn}, and incremental network quantization (INQ)~\cite{inq}, but we are focusing on binary network in this paper. CNNs with low-precision weights have been shown to exhibit extremely closed performance as their real-value counterparts, on semantic-level tasks such as image classification~\cite{BinaryConnect} (like TWN, XNOR, QNN, INQ, etc.); however, it is widely presumed that CNNs with low-precision weights would fail on pixel-level tasks such as image reconstruction and super resolution, due to the reduced model complexity.

In this work, we are proposing an efficient network binarization strategy for image super-resolution tasks, based on BinaryNet and XnorNet~\cite{Courbariaux,xnor}, to speed up inference and simultaneously achieve similar image quality as the full-precision network. To verify the effectiveness of the proposed strategy for general CNN-based SR methods, we evaluate two state-of-the-art SR models SRGAN~\cite{sisrgan} and LapSRN~\cite{Lai}, and compare them with their real-value counterparts, on three benchmark metrics the peak-signal-to-noise-ratio (PSNR), structure similarity (SSIM)~\cite{metrics}, and information fidelity criterion (IFC) calculated on $y$-channel to evaluate the performance of super resolution.

\section{SR Network Binarization}
\label{sec:net_binarization}

The original binary neural networks are designed and applied to semantic-level tasks like classification~\cite{Courbariaux,xnor}, without sacrificing much accuracy.
Unlike semantic-level tasks, SR is an image reconstruction task that aims at pixel-level accuracy.
SR networks with full binarization in nature would suffer from poor performance, because limited parameters restrain representation power of the network and therefore affect reconstruction results.
In Figure~\ref{ful_binary}, we show the results generated from fully binarized SRResNet network~\cite{sisrgan}, with a upscaling factor of $4$.
From the results, we can see that fully binarized SR network significantly decreases the color richness.
Therefore, network binarization needs to be specially handled for SR network, as well as other image reconstruction tasks.

\subsection{Motivation}
\label{sec:motivation}
For a neural network, deeper network structure usually means more powerful representation ability, and therefore better performance, but in higher difficulty of convergence.
Residual blocks are introduced in~\cite{HeZRS15} to facilitate the training process of very deep neural networks for image classification tasks. 
The structure enables network to learn the identity function efficiently.
It is also an appealing property for SR networks, since the output HR image share basic color and structure with the input LR image. 
Recent SR algorithms utilize residual structures in their deep neural networks and achieve decent results. 
Johnson et al.~\cite{JohnsonAL16} and Ledig et al.~\cite{sisrgan} used five and sixteen residual blocks in their SR neural networks, respectively.

Image pyramid is a multi-scale representation technique that is commonly used for image reconstruction tasks~\cite{adelson1984pyramid,burt1987laplacian}.
Using the pyramid technique, the images can be represented as a series of band-pass filtered images.
The base layer represents the low-frequency information and the gradient layers captures  sub-band information and most high-frequency details.
Similarly, SR tasks can be considered as reconstructing high-frequency information based on a low-pass filtered image (LR image).
And the recent residual-structure-based SR networks function in a similar way as the image-pyramid reconstruction process, where the residual structure and the skip connection in neural network resemble to the gradient layers and the upsampled images from coarser scales in image pyramid.

\begin{figure}[h]
\begin{center}
\begin{tabular}{c}
\includegraphics[width=0.8\linewidth]{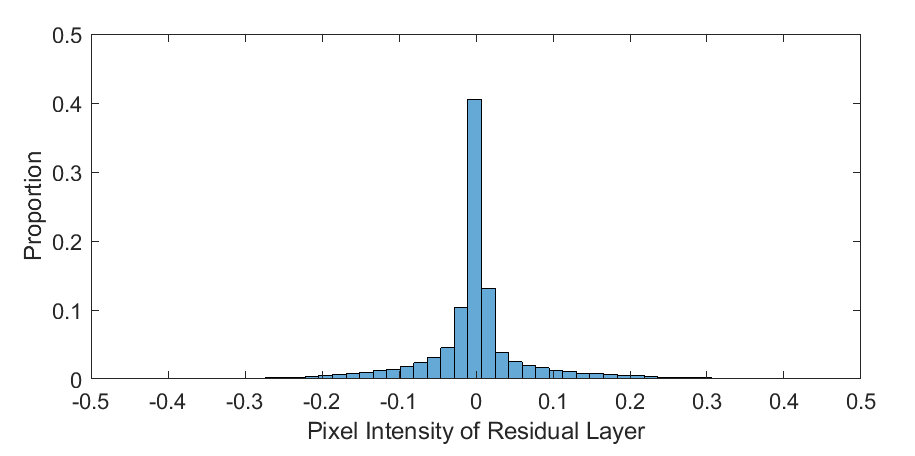}
\end{tabular}
\caption{Statistical results on histograms of the gradient layers of Laplacian pyramid. The range of the image pixel is $[0, 1]$.}
\label{lapgrad}
\end{center}
\end{figure}

In image pyramid, the gradient layers are known to hold good sparsity property.
We study the histograms of gradient layers of a Laplacian pyramid~\cite{burt1987laplacian}, 
and show the results in Figure~\ref{lapgrad}.
From the figure, we can clearly see that 1) very few pixels have values (sparsity) and 2) value distribution is highly centralized around $0$.
Those facts indicate that the gradient layer of a Laplacian pyramid can be approximated by a layer with a small number of values, and therefore motivate us to represent it via binary filters with corresponding scaling filters in a neural network.

\subsection{Binarization Method}
Binary weights, from its quantization property in nature, would cause information loss comparing to using real weights. 
Since SR algorithms aim at enhancing image details from its original image, binary weights should be carefully used in the network.
Based on the analysis in Section~\ref{sec:motivation}, we propose a SR network binarization strategy that binarizes the residual blocks and couples each binary filter with a learnable weight in the network.
The convolutional filters outside the residual blocks will be kept in real-values, and the propagation and parameter update steps are the same as usual.

\begin{figure*}[ht]
\centering
\includegraphics[width=0.98\textwidth]{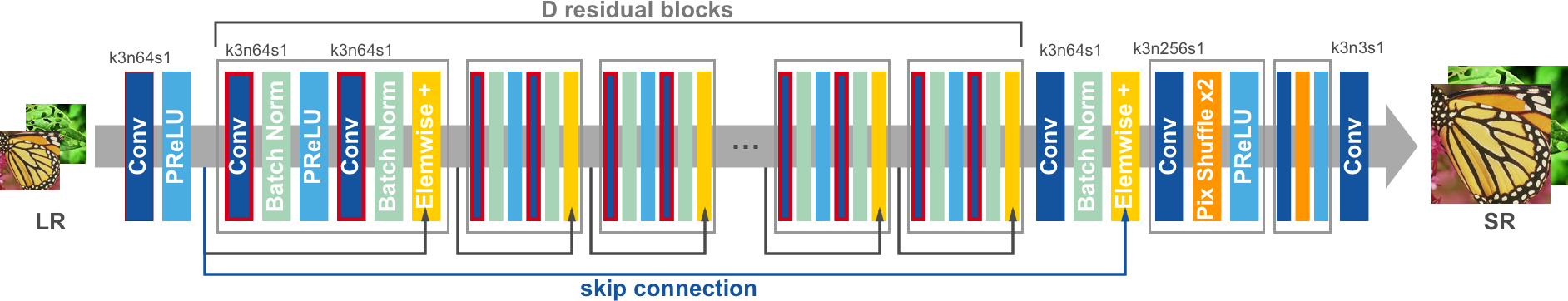}
\caption{System architecture of the generator network in SRResNet/SRGAN~\cite{sisrgan} that transforms the low-resolution input image $I^{LR}$ to high-resolution image $I^{SR}$. The red boxes indicate the binarized layers in our network. $k$ is the filter size, $n$ denotes the number of filters and $s$ represents the stride number in convolution layer.}
\label{generator}
\end{figure*}

\begin{figure*}[ht]
\centering
\includegraphics[width=0.95\textwidth]{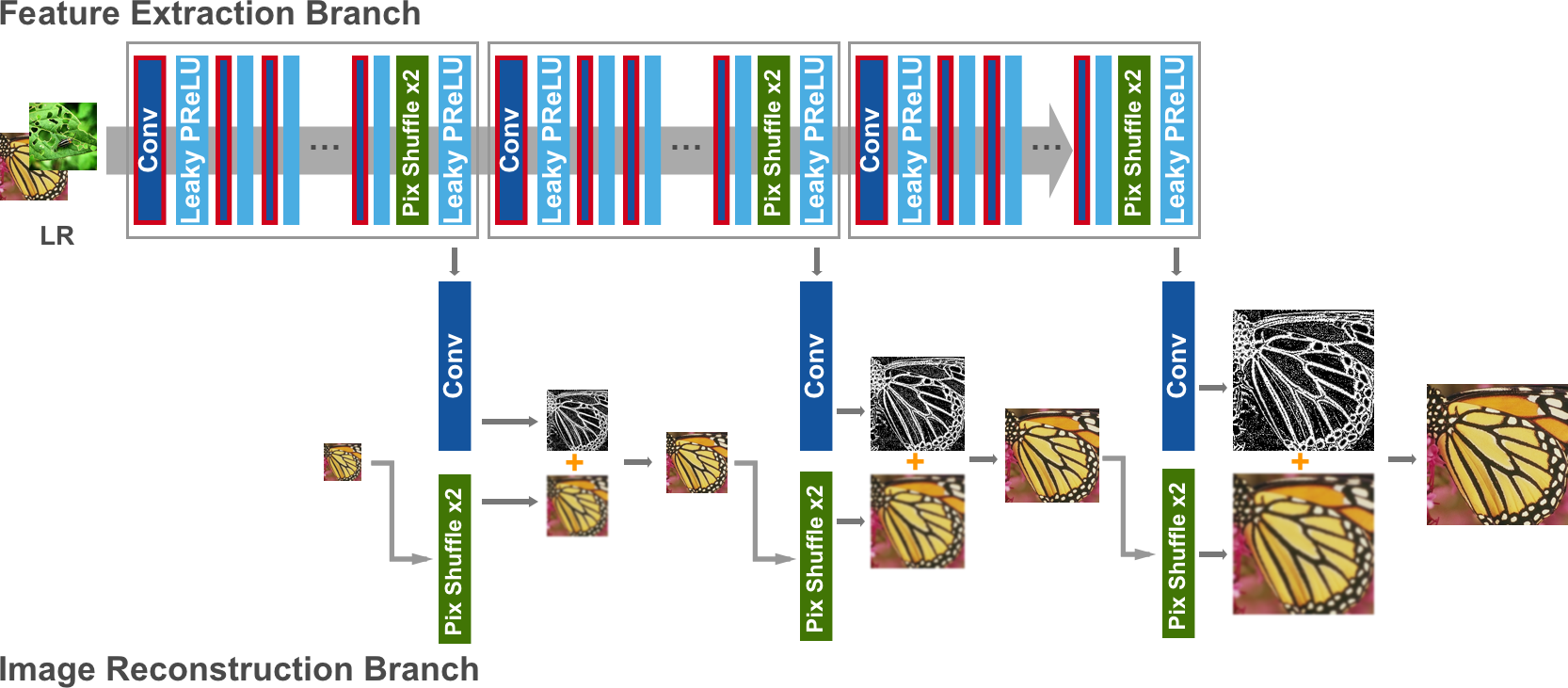}
\caption{System architecture of the Laplacian Pyramid network~\cite{Lai} for SR. The red boxes indicate binarized layers in our network. }
\label{lap}
\end{figure*}

\begin{figure*}[ht]
\centering
\includegraphics[width=0.95\textwidth]{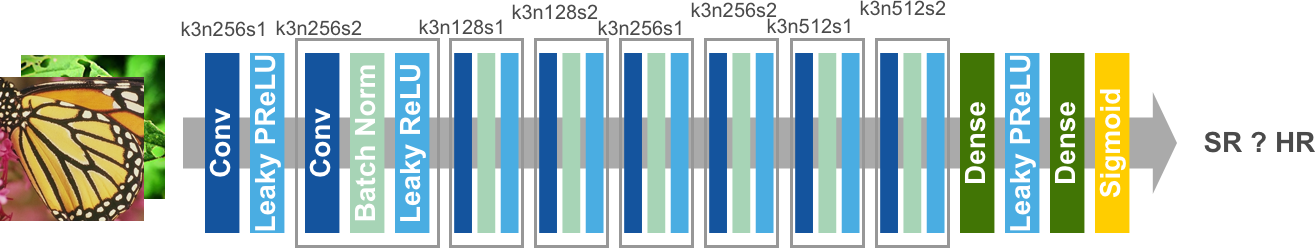}
\caption{System architecture of the discriminator network in SRGAN~\cite{sisrgan} that is trained to distinguish super-resolution results $I^{SR}$ from high-resolution image $I^{HR}$. $k$ is the filter size, $n$ denotes the number of filters and $s$ represents the stride number in convolution layer.}
\label{discriminator}
\end{figure*}

To binarize residual blocks in the network, we would like to approximate real-value weight filters in residual blocks with binary-value weight filters and scaling factors.
We first define our binarization function~\cite{Courbariaux} to transform real values to $+1$ or $-1$:
\begin{equation}
x^b = \textrm{sign}(x) \left\{
    \begin{array}{ll}
    +1, \textnormal{if } x \geqslant 0,\\
    -1, \textnormal{otherwise},\\
    \end{array}
    \right.
\label{binarization}
\end{equation}
where $x$ is the real-valued variable, and $x^b$ is the corresponding binarized weight. For the practical applications, booleans can be used to represent the binarized weights.
In the binarized network, a scaling factor $\alpha$ is assigned to each binarized filter, so a real-value weight filter $W$ in the convolutional neural network is replaced with $\alpha \textrm{sign}(W)$. The factor $\alpha$ is a vector of the same length as the output channel size of the corresponding convolutional layer.

Each iteration of the training phase takes pairs of low-resolution image patches $I^{LR}$ and corresponding high-resolution image patches $I^{HR}$, cost function $C(I^{HR},I^{SR})$ and learning rate $\eta$ as inputs, and updates weights $W^{t}$ as outputs.
For the forward propagation, we first update factor $\alpha$ and binarize the real-value weight $B = \textrm{sign}(W)$, then $\alpha B$ is used for the following computation.
Algorithm~\ref{BNN forward} demonstrates the procedure of forward propagation, and we will describe $\alpha$ updating strategy in Section~\ref{sec:factor_update}.
For the backward propagation as shown in Algorithm~\ref{BNN backward}, the gradients of the weights are kept in real values. Similar to~\cite{xnor}, we take derivatives of $l$-th layer cost $C_l$ with respect to the binarized weight $B_{lk}$ as $\dfrac{\partial C_l}{\partial B_{lk}}$. The gradients are clipped to range $(-5,5)$ for stablity, and are used to update the real weight by $\dfrac{1}{\alpha_{lk}} \dfrac{\partial C_l}{\partial B_{lk}}$.
The real-value parameters and binary parameters are then updated by accumulating gradients as shown in Algorithm~\ref{Accumulating gradients of parameters}.

\begin{algorithm}
\caption{Binary Residual Block Forward Propagation}
\begin{algorithmic}
\FOR{$l=1$ to $L$ layer}
\FOR{$k=1$ to $K$ output channel}
\STATE $\alpha_{lk} \leftarrow \textrm{AlphaUpdate}()$  {\footnotesize //update alpha according to \eqref{learn alpha} or \eqref{determ alpha}}
\STATE $B_{lk} \leftarrow \textrm{sign}(W_{lk})$
\ENDFOR
\ENDFOR
\STATE $I^{SR} \leftarrow \textrm{BinaryForward}(I^{LR}, \alpha B)$
\end{algorithmic}
\label{BNN forward}
\end{algorithm}

\begin{algorithm}
\caption{Binary Residual Block Backward Propagation}
\begin{algorithmic}
\FOR {$l=L$ to $1$}
\FOR{$k=1$ to $K$ output channel}
\STATE $g_{W_{lk}} \leftarrow \textrm{Clip}(\dfrac{1}{\alpha_{lk}} \dfrac{\partial C_l}{\partial B_{lk}}, -5, 5)$
\STATE $W_{lk} \leftarrow \textrm{BinaryParameterUpdate}(W_{lk}, \eta, g_{W_{lk}})$
\ENDFOR
\ENDFOR
\end{algorithmic}
\label{BNN backward}
\end{algorithm}

\begin{algorithm}
\caption{Accumulating gradients of parameters}
\begin{algorithmic}
\FOR{$t=0$ to $T$}
\STATE $\theta^{t+1} \leftarrow \textrm{ParameterUpdate}(\theta^{t}, \eta^{t}, g_{\theta})$  {\footnotesize//update non-binary parameters}
\STATE $W^{t+1} \leftarrow \textrm{BinaryParameterUpdate}(W^{t}, \eta^{t}, g_{W})$  {\footnotesize//update binary parameters}
\STATE $\eta^{t+1} \leftarrow \textrm{LearningRateUpdate}(\eta^{t})$\\
\ENDFOR
\end{algorithmic}
\label{Accumulating gradients of parameters}
\end{algorithm}

\subsection{Scaling Factors for Binary Filters}
\label{sec:factor_update}
In~\cite{xnor}, the scaling vector $\alpha$ is determined in a deterministic way, by solving the following optimization:
\begin{equation}
J(B,\alpha) = ||W - \alpha B||^2,
\end{equation}
where binary filter $B=\textrm{sign}(W)$ is obtained by the binarization function.
The optimal value for the scaling factor $\alpha \ast$ is calculated as:
\begin{equation}
\alpha^{\ast} = \frac{1}{n} ||W||_{l1}.
\end{equation}
However, this $\alpha^{\ast}$ is the optimal solution for estimating $W$ with $\alpha B$, which is not our global objective for the SR problem. 

To better approximate the gradient information, we propose to train $\alpha$ as a parameter in the network, 
\begin{equation}
\alpha_{c}^{t} \leftarrow \textrm{ParameterUpdate}(\alpha_{c}^{t-1}, \eta, g_{\alpha_{c}}),
\label{learn alpha}
\end{equation}
instead of
\begin{equation}
\alpha_{c} \leftarrow \textrm{avg}(\textrm{abs}|W_{c}|_{l1}).
\label{determ alpha}
\end{equation}
We will provide comparisons between these two methods of scaling factor update in Section~\ref{sec:experiment}.

\section{Experiment}
\label{sec:experiment}

To verify the proposed binarization strategy for SR tasks, we evaluate this strategy on several state-of-the-art SR architectures: SRResNet/SRGAN~\cite{sisrgan} and LapSRN~\cite{Lai}.
We evaluate the methods in NTIRE 2017 dataset~\cite{ntire}, which includes $800$ HR training images and $100$ HR validation images as our training and test datasets. 
To train/test $2\times$ and $4\times$ models, the corresponding LR images are generated using the bicubic downsampling, with a scale factor of $2\times$ and $4\times$, respectively. 

\subsection{Training Details}
We test our binarized neural network on three different models, namely SRResNet, SRGAN, and LapSRN, all of which employ residual blocks in their networks. 
We will describe the models and training details for each method below. For all the three models, we set larger learning rates for the binary version, with a factor of $3\times$-$4\times$ comparing to those of the real-value network. The reason is that it requires larger momentum to change the binary weights or switching signs. Within this learning rate setting, the binary network converges at a similar pace as the real-value network. We show the learning curves for both real-value networks and their binary weight counterparts in the supplementary material.
We use a batch size of $16$ for all the training.
For data augmentation, we adopt the following approaches: 1) randomly cropping patches; 2) randomly flipping horizontally or vertically, rotation and noise; 3) randomly rotating images by $\{0,90,180,270\}$ degrees; 4) adding Gaussian noise to HR training patches.

\textbf{SRResNet} We provide the binarized SRResNet structure in Figure~\ref{generator} and show the binarized layers using red boxes. Basically, we binarize all the convolutional layers in the residual blocks, using the algorithms~\ref{BNN forward}, \ref{BNN backward} and~\ref{Accumulating gradients of parameters}. 
We train real-weight and binary-weight SRResNets for $500$ epochs, with each epoch containing $50$ iterations. We use $1\times10^{-4}$ and $3\times10^{-4}$ respectively, as our initial learning rate, and a decay of $0.9$ in every $20$ epochs.


\textbf{SRGAN} The same method of binarization as SRResNet is used in the generator of SRGAN. We keep the discriminator in real weights, as the purpose of binarization is to improve inference performance in terms of speed and model storage size and the discriminator does not affect inference performance. 
We pretrain real-weight and binary-weight generators of SRGAN for $100$ epochs with learning rates of $1\times10^{-4}$ and $3\times10^{-4}$ respectively. Then we jointly train generator and discriminator for $500$ epochs with a learning rate of $1\times10^{-4}$ as our initial learning rate, and a decay of $0.9$ in every $20$ epochs.


\textbf{LapSRN} The Laplacian Pyramid network proposed by Lai et al.~\cite{Lai} consists of $D$ residual blocks in the Feature Extraction Branch. We binarized all convolutional layers in residual blocks, i.e. all weights of convolutional filters in the branch, as shown in Figure~\ref{lap}.
We train real-weight and binary-weight LapSRN for $300$ epochs, with each epoch containing $800$ iterations. We use $3\times10^{-5}$ and $1\times10^{-4}$ respectively, as the initial learning rates, and a decay of $0.8$ in every $30$ epochs.


\begin{figure*}[t]
\begin{center}
\begin{tabular}{cccc}
\includegraphics[width=0.18\linewidth]{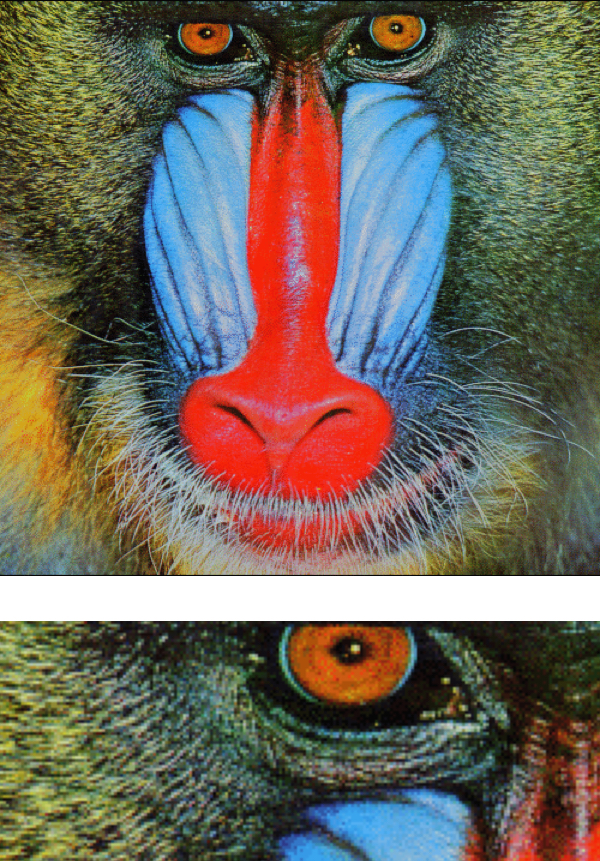} & \hspace{-1mm}
\includegraphics[width=0.18\linewidth]{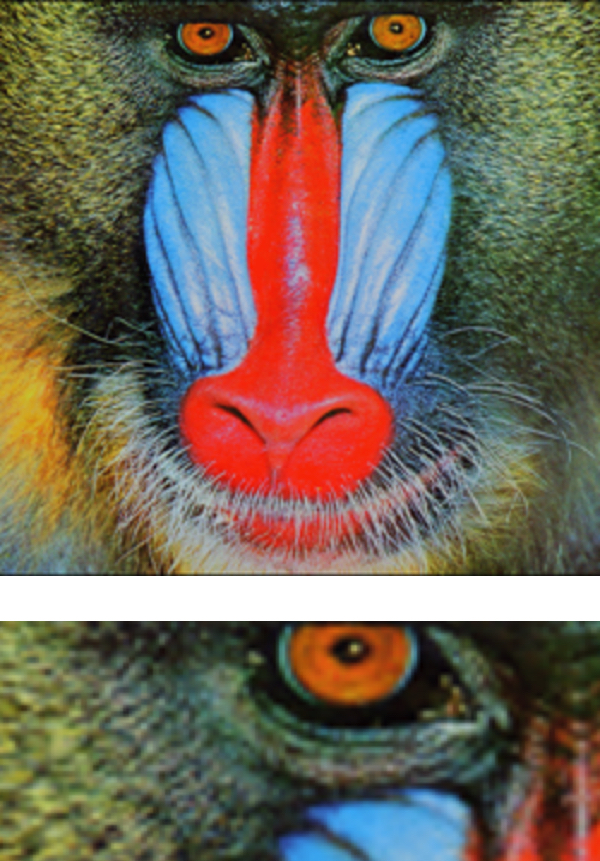} & \hspace{-1mm}
\includegraphics[width=0.18\linewidth]{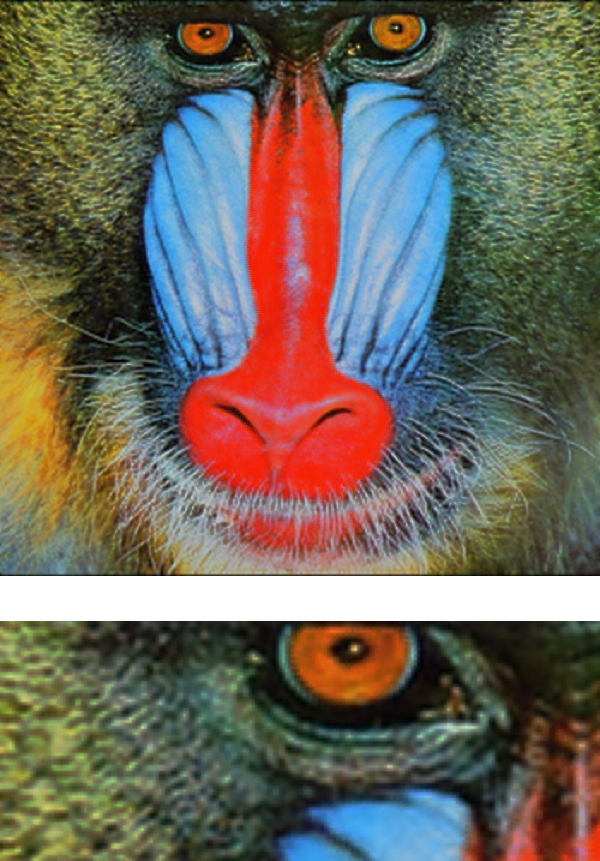} & \hspace{-1mm}
\includegraphics[width=0.18\linewidth]{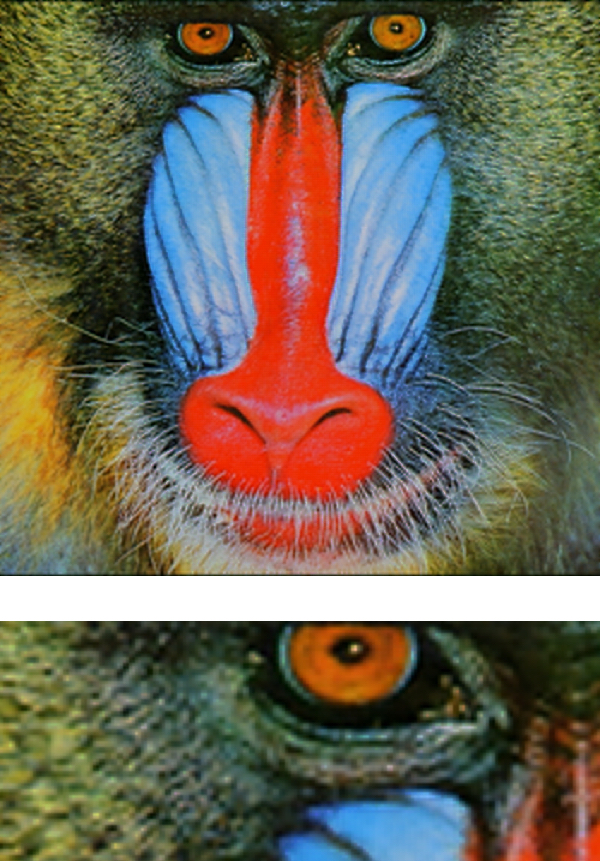} \\
Ground-truth & Bicubic & Real-weight LapSRN& Binary-weight LapSRN\\
\includegraphics[width=0.18\linewidth]{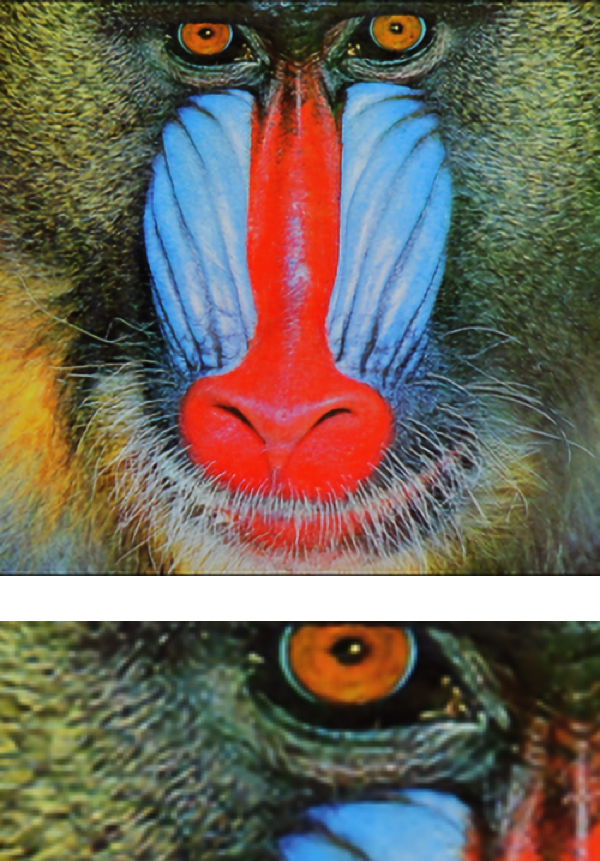} & \hspace{-1mm}
\includegraphics[width=0.18\linewidth]{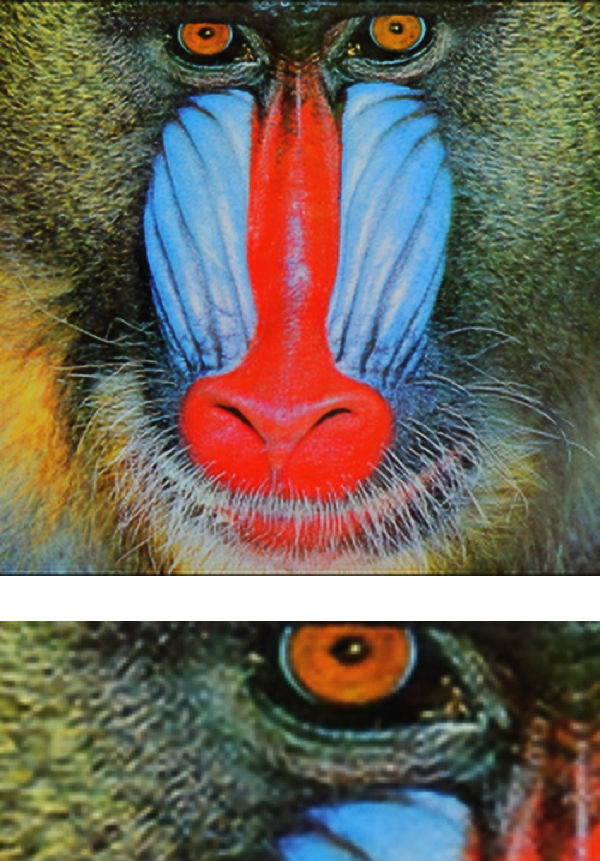} & \hspace{-1mm}
\includegraphics[width=0.18\linewidth]{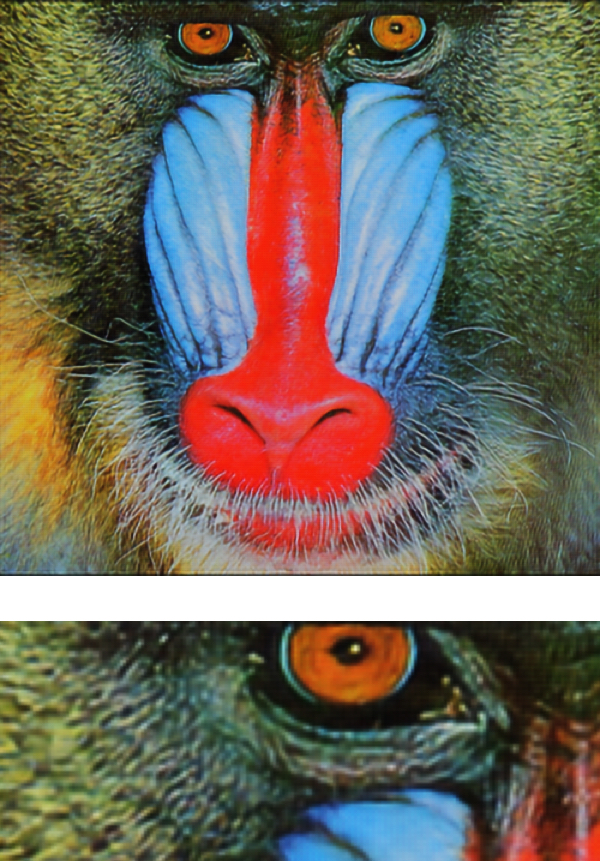} & \hspace{-1mm}
\includegraphics[width=0.18\linewidth]{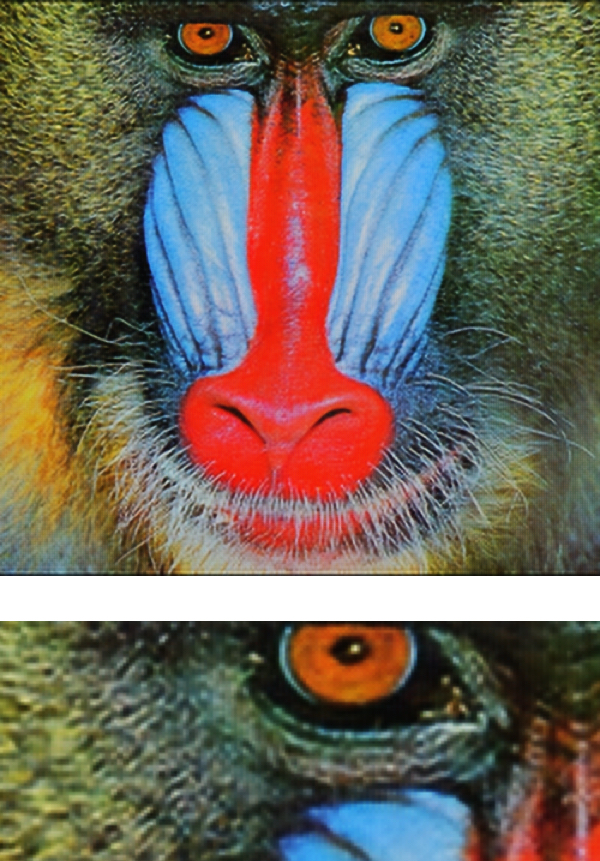} \\
Real-weight SRResNet & Binary-weight SRResNet & Real-weight SRGAN& Binary-weight SRGAN\\
\end{tabular}
\caption{Comparisons between real-weight networks and their binarized versions with an upsampling factor of $2$.}
\label{2x}
\end{center}
\end{figure*}

\begin{figure*}[t]
\begin{center}
\begin{tabular}{cccc}
\includegraphics[width=0.18\linewidth]{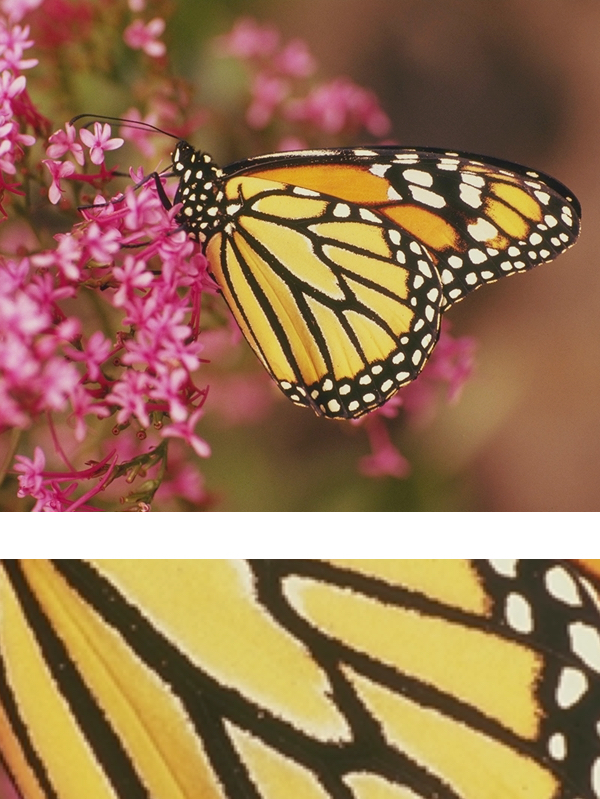} & \hspace{-1mm}
\includegraphics[width=0.18\linewidth]{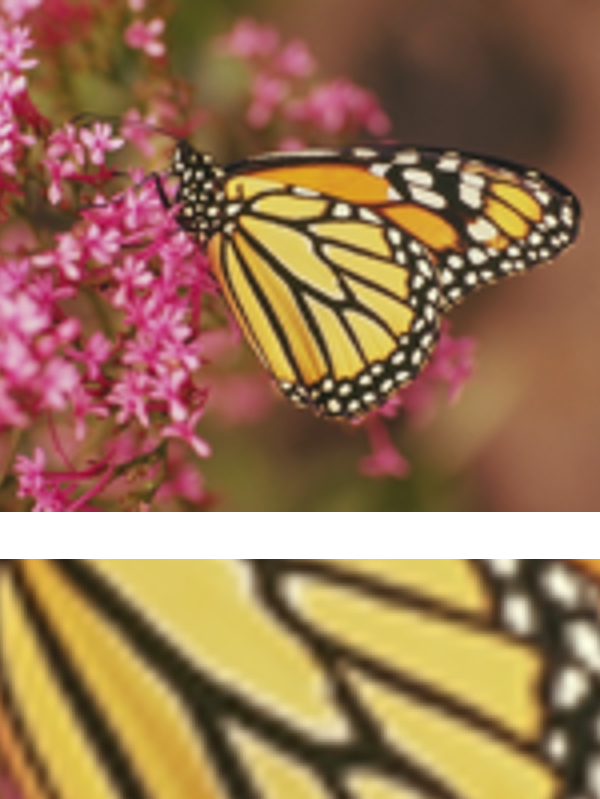} & \hspace{-1mm}
\includegraphics[width=0.18\linewidth]{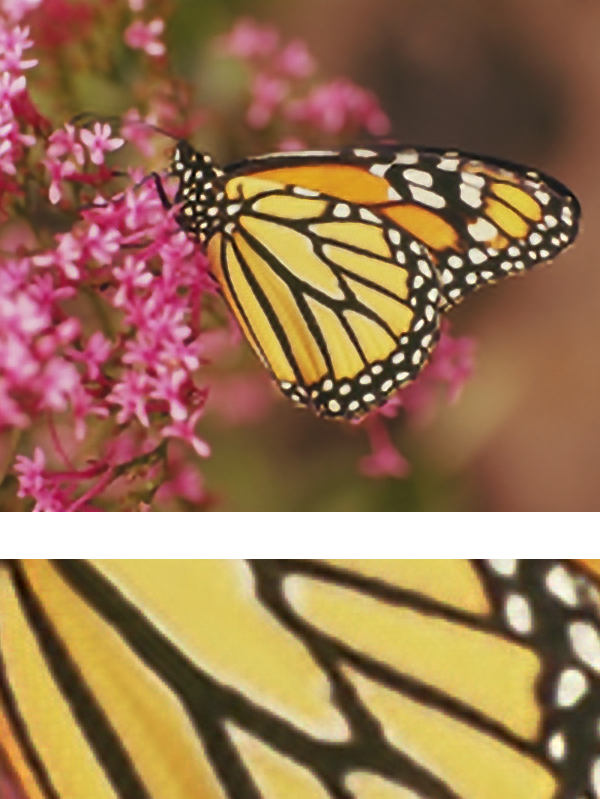} & \hspace{-1mm}
\includegraphics[width=0.18\linewidth]{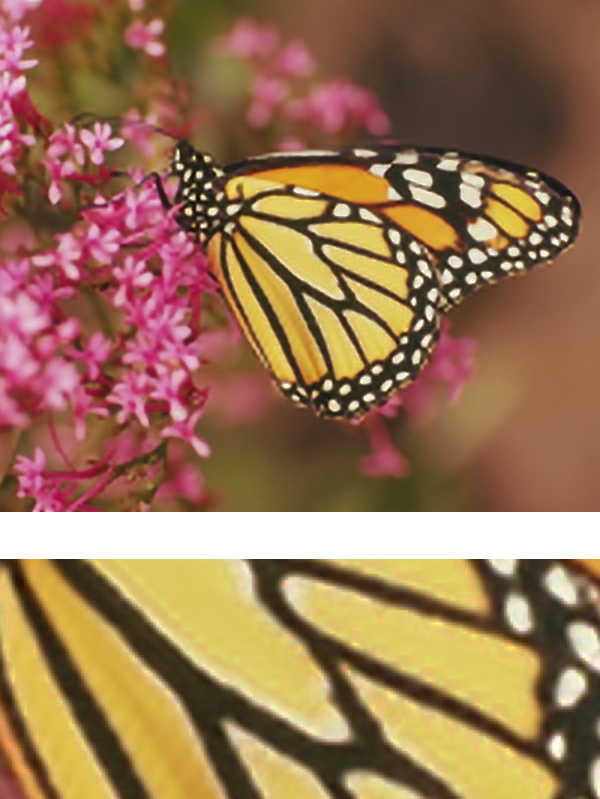} \\
Ground-truth & Bicubic & Real-weight LapSRN & Binary-weight LapSRN\\
\includegraphics[width=0.18\linewidth]{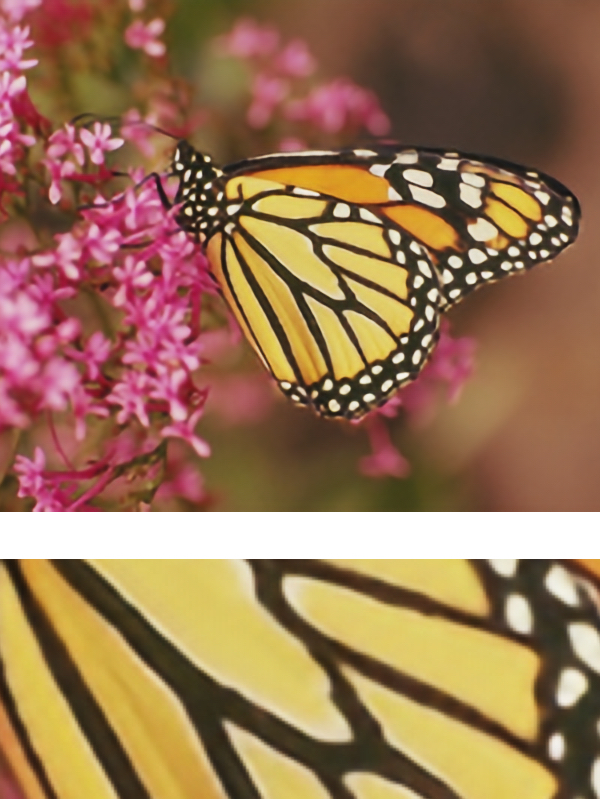} & \hspace{-1mm}
\includegraphics[width=0.18\linewidth]{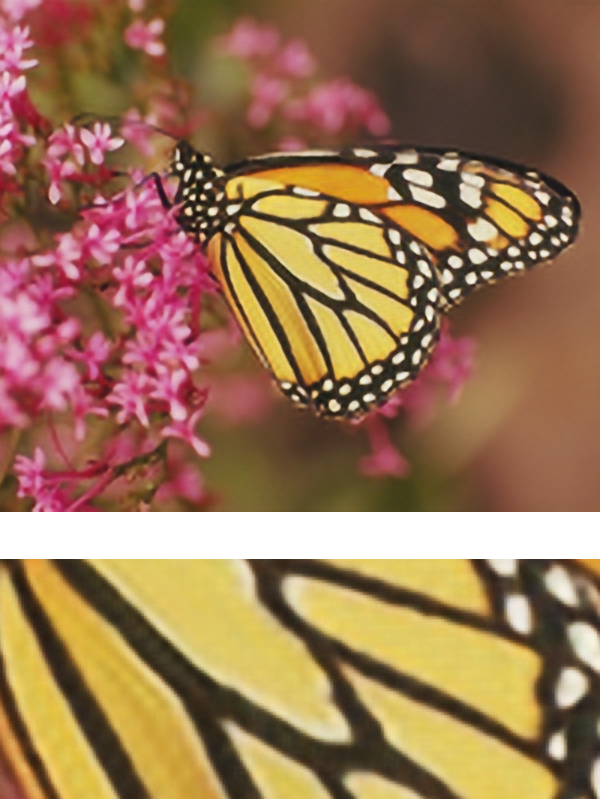} & \hspace{-1mm}
\includegraphics[width=0.18\linewidth]{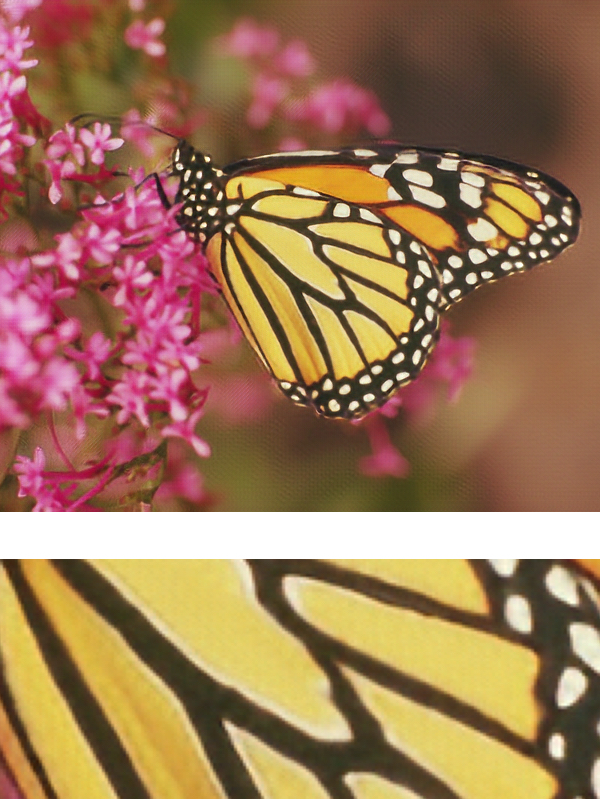} & \hspace{-1mm}
\includegraphics[width=0.18\linewidth]{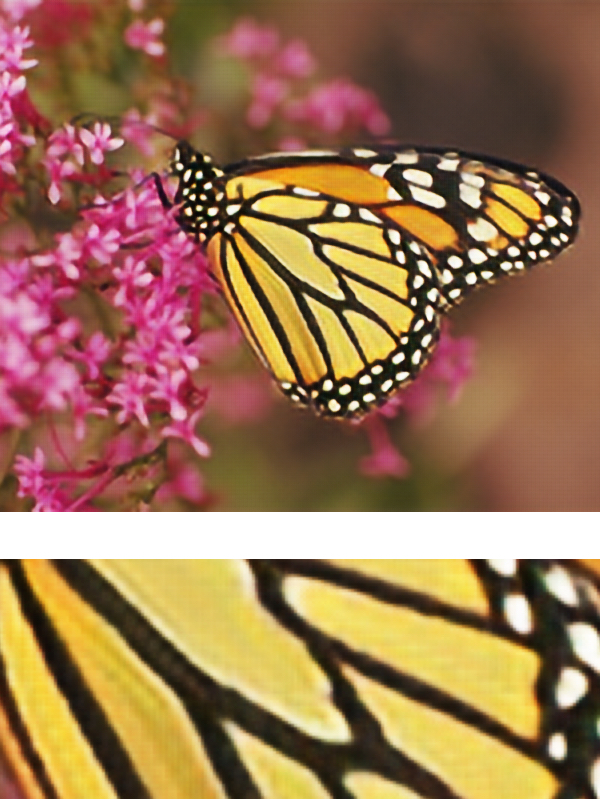} \\
Real-weight SRResNet & Binary-weight SRResNet & Real-weight SRGAN& Binary-weight SRGAN\\
\end{tabular}
\caption{Comparisons between real-weight networks and their binarized versions with an upsampling factor of $4$.}
\label{4x}
\end{center}
\end{figure*}

\begin{table*}[ht]
\centering
\resizebox{\textwidth}{!}{%
\begin{tabular}{l|cllc}
\hline
\multicolumn{1}{c}{\multirow{2}{*}{\textbf{Algorithm}}} & \multirow{2}{*}{\textbf{Scale}} & \multicolumn{1}{c}{\textbf{Set5}} & \multicolumn{1}{c}{\textbf{Set14}} & \multicolumn{1}{c}{\textbf{Urban100}} \\
\multicolumn{1}{c}{}                                    &                                 & \multicolumn{1}{c}{PSNR/SSIM/IFC} & \multicolumn{1}{c}{PSNR/SSIM/IFC}  &  \multicolumn{1}{c}{PSNR/SSIM/IFC}     \\ \hline
\hline
Bicubic & 4 &28.43 / 0.811 / 2.337 &25.90 / 0.704 / 2.246  & 23.15 / 0.660 / 2.386 \\
\hline
\textbf{SRResNet}~\cite{sisrgan} & 4 & 31.10 / 0.875 / 3.091 & 27.59 / 0.764 / 2.743
&25.09 / 0.750 / 3.040
\\
\textbf{SRResNet Binary (ours)} & 4 & 30.34 / 0.864 / 3.052 & 27.16 / 0.756 / 2.749
& 24.48 / 0.728 / 2.913
\\
\hline
\textbf{SRGAN}~\cite{sisrgan} & 4 & 30.43 / 0.855 / 2.862 & 27.00 / 0.749 / 2.493
 & 24.75 / 0.734 / 2.865
\\
\textbf{SRGAN Binary (ours)} & 4 & 
30.13 /	0.853 / 2.547 & 26.98 / 0.746 / 2.336 &24.31 / 0.716 / 2.499
\\
\hline
\textbf{LapSRN (4x)~\cite{Lai}} & 4 & 30.28 / 0.858 / 2.851 &27.15 / 0.748 / 2.566
 &24.37 / 0.722 / 2.755
\\
\textbf{LapSRN Binary (4x) (ours)} & 4 & 30.21 / 0.857 / 2.865& 27.13 / 0.751 / 2.616 & 24.31 / 0.720 / 2.735\\
\hline
\hline
Bicubic & 2 &33.69 / 0.931 / 6.166 & 30.25 / 0.870 / 6.126 &26.89 / 0.841 / 6.319\\
\hline
\textbf{SRResNet}~\cite{sisrgan} & 2 & 36.36 / 0.952 / 6.714 & 32.16 /	0.904 / 6.577 
 &29.96 / 0.901 / 7.016
 \\
\textbf{SRResNet Binary (ours)} & 2 & 35.66 / 0.946 / 5.936 & 31.56 /	0.897 /	5.893 &
 28.76 / 0.882 / 6.112
\\
\hline
\textbf{SRGAN}~\cite{sisrgan} & 2 & 35.31 / 0.941 / 6.332 & 31.81 / 0.901 / 6.280&
 29.63 / 0.897 / 6.646
\\
\textbf{SRGAN Binary (ours)} & 2 & 34.91 / 0.938 / 5.840 & 30.92 / 0.892 / 6.354 & 28.55 / 0.878 / 6.490
\\
\hline
\end{tabular}}
\caption{Quantitative evaluation of state-of-the-art SR algorithms and their binarized versions.}
\label{metrics}
\end{table*}

\subsection{Evaluating Results}
We evaluate our models on three widely used benchmark datasets \textbf{Set5}~\cite{bevilacqua2012low}, \textbf{Set14}~\cite{zeyde2010single}, and \textbf{Urban100}~\cite{urban100}. To evaluate our method, we provide comparisons of the SR images from the following models and their corresponding binarized versions: SRResNet~\cite{sisrgan}, SRGAN~\cite{sisrgan}, and LapSRN~\cite{Lai}. The bicubic upsampling method is used as a baseline. 

From the results in Figure~\ref{2x} and~\ref{4x}, we can see that the binarized networks using the proposed method perform similarly to their real-weight counterpart.
Although our binarized networks achieves comparable results as their real-weight counterparts, slightly more aliasing artifacts have been introduced in SRResNet $2\times$ and $4\times$ if watching closely.
That is, the binarization strategy still compromises performance in terms of high-frequency details.
So how to further improve the performance of binary network will be our future work. 

More experiment results can be found in supplementary material.

\begin{table}[ht]
\begin{center}
\begin{tabular}{| c | c | c |}
 \hline
 Model (4x) & deterministic & learnable\\
 \hline
SRResNet & 27.06/0.754/2.712 & \textbf{27.16}/\textbf{0.756}/\textbf{2.749} \\
 \hline
LapSRN & 27.02/0.747/2.604 & \textbf{27.13}/\textbf{0.751}/\textbf{2.616} \\
 \hline
\end{tabular}
\caption{Comparison on networks using deterministic and learnable scaling factors. The results are evaluated on SRResNet~\cite{sisrgan} and LapSRN~\cite{Lai} structures in PSNR/SSIM/IFC.}
\label{deter_vs_learn}
\end{center}
\end{table}

For quantitative evaluation, we calculate PSNR, SSIM and IFC~\cite{metrics} measures between generated images and the ground truths, and all the reported values in Table~\ref{metrics} are calculated on the Y-channel of YUV color space. Both the real-weight and the binary-weight networks achieve better results than bicubic baseline in terms of the metrics. In most of the cases, the binary-weight networks perform similarly with their real-weight counterparts, with a difference less than 0.5 PSNR or 0.005 SSIM.
And the margin becomes even smaller for upsampling facotr of $4$.

\begin{table}[ht]
\begin{center}
\begin{tabular}{| c | c | c | c |}
 \hline
 Num. of Res-Blocks & PSNR & SSIM & IFC\\
 \hline
 8 & 26.74& 0.738 & 2.136 \\
 \hline
 16 & 27.16 & \textbf{0.756} & 2.749 \\
 \hline
 24 & \textbf{27.19}& 0.754 & \textbf{2.774} \\
 \hline
\end{tabular}
\caption{Performance of binarized SRRestNet (4x) with 8, 16 and 24 residual blocks on Set14~\cite{zeyde2010single} dataset.}
\label{resblock}
\end{center}
\end{table}


\subsection{Model Analysis}
To verify the effectiveness of the proposed learnable scaling factor for binary filters, we conduct experiments and show the comparison between networks using the learnable scaling factor and the deterministic one in Table~\ref{deter_vs_learn}. In general, the networks using learnable scaling factor have faster convergence and achieve better results comparing to the deterministic ones.

\begin{table*}[ht]
\centering
\resizebox{\linewidth}{!}{
\begin{tabular}{c|cccc}
\hline
\multirow{2}{*}{Algorithm} & \# of Res-Block / & \# of Binary Params & \# of Real Params & Model Size \\
& Res-Conv-Layer& Binary/Real Network & Binary/Real Network & Binary/Real Network\\ \hline
\hline
\textbf{SRResNet (2x)}& 16 / 32 & 1,179,648 / 0& 197,059 / 1,374,659  &  0.928 / 5.499MB\\ \hline 
\textbf{SRResNet (4x)}& 8 / 16 & 589,824 / 0& 339,651 / 928,451 & 1.428 / 3.714MB   \\ \hline
\textbf{SRResNet (4x)}& 16 / 32& 1,179,648 / 0& 344,771 / 1,522,371 &1.518 / 6.089MB  \\ \hline 
\textbf{SRResNet (4x)}& 24 / 48&1,769,472 / 0& 349,801 / 2,116,201 &1.608 / 8.465MB  \\ \hline 
\textbf{LapSRN (4x)}& 10 / 10 &368,640 / 0& 152,656 / 520,656&1.494 / 2.083MB   \\ \hline 
\end{tabular}}
\caption{Number of parameters and model size of binary and real-weight networks (32-bit floating point). The parameters of binary networks include binary parameters of filters and float parameters assigned to filters.}
\label{mem}
\end{table*}

The binarized networks reduce the computational load and enable efficient inference.
As the deeper networks have better representation abilities and usually generate better results, we also evaluate the depth of the binarized networks, all of which enable efficient inference.
We train SRResNet-Binary (4x) with $8$, $16$ and $24$ residual blocks, and show the quantitative comparisons in Table~\ref{resblock}.
We note that we use $16$ residual blocks for the SRResNet in all other experiments, as the same as used in~\cite{sisrgan}.
From the results, we can see that deeper networks perform better than shallower ones in general. 
However, marginal improvement is achieved after using more residual blocks over $16$, and this also suggests that binarized layers in $16$ residual blocks are sufficient to represent the high-frequency information of SR images, which also verifies the design of the proposed binarization strategy.

\begin{table}[ht]
\begin{center}
\begin{tabular}{| c | c | c |}
 \hline
 Vote Percentage & $2\times$ & $4\times$ \\
 \hline
 SRResNet & 0.50 & 0.45 \\
 SRResNet-Bin & 0.50 & 0.55 \\
 \hline
 SRGAN & 0.60 & 0.62 \\
 SRGAN-Bin & 0.40 & 0.38 \\
 \hline
 LapSRN & 0.51 & 0.55 \\
 LapSRN-Bin & 0.49 & 0.45 \\
  \hline 
\end{tabular}
\caption{User study for $2\times$ and $4\times$ SR results from binary and real-weight networks. The test images are randomly selected from \textbf{Set5}~\cite{bevilacqua2012low} and \textbf{Set14}~\cite{zeyde2010single}.}
\vspace{-3mm}
\label{user}
\end{center}
\end{table}

\subsection{Model Size and Computational Complexity}
In Table ~\ref{mem}, we list the number of parameters and model size of the networks that we test. In TensorFlow, we use $32$ or $64$ bits floating point precision to represents real parameters, and the parameter binarization would reduce it to a single bit, which is a factor of $32$ or $64$ in terms of memory used for model storage. 
Even though this binarization strategy only applies to residual blocks in the network, it still yields a $50\%-80\%$ model compression depending on the network structure. Higher portion of residual components would yield higher compression rate. This also provides a network design guideline for applications that have memory limitation, e.g., on-device processing in IoT edge devices and mobile platforms. 

Another benefit of network binarization is to reduce computational complexity during inference, as binarized filter weights enables bit operations instead of float multiplication (flops) when computing convolution.
For a single convolution layer with input tensor size $(M,N,C)$ and convolutional filters of size $(K,K,C,F)$, there are in total $O(MNC^2K^2F)$ multiplication floating-operations (flops), which will be replaced with bit operations after binarization.
Modern computers can compute 4 flops or 512 bit operations per clock cycle, and the op of addition is $\sim 3\times$ faster than that of multiplication, 
For a SRResNet 4x model with 16 residual blocks on an image of size $1200\times800\times3$, there are in total $\sim 10^{14}$ flops, with about the same number of flops of multiplication and addition. 
In this model, more than $75\%$ of multiplication flops are replaced with bit operations after binarization.
Thus, there is a potential speedup of $\sim 2\times$ for this SRResNet(4x) model.
Applying same calculation to SRResNet(2x) model would result in a $\sim 5\times$ computational gain.


\subsection{User Study}
To better evaluate the visual quality of the results from real-weight network and its binarized version, we conducted a user study on the SR results. 
We develop a web-based system to display and collect study results. 
The system provides two side-by-side results at a time, one from the real-weight network with scaling factor $2\times$ or $4\times$, and another from its binary counterpart.
Each pair of results is randomly selected and placed from the dataset \textbf{Set5}~\cite{bevilacqua2012low} and \textbf{Set14}~\cite{zeyde2010single}.
We have collected results of their preferred images from 24 users, each user is asked to rate $20$ pairs of images. 
The results are shown in Table~\ref{user}, and we can see our binarized SR models perform similarly as real-weight SR models.




\section{Limitation and Future Work}
The proposed binarization strategy are designed for SR tasks, but it is possible to apply this strategy to other pixel-level tasks like denoising and deblurring. 
However, specific network designs are needed when applying binarization to other tasks, e.g., an kernel estimation process in deblurring may require special handling. And how to apply the binarization strategy to other pixel-level image reconstruction tasks could be a good future research direction.

Although we are able to replace the float multiplications with bit-wise operation when computing convolution, there are still rooms of efficiency improvement using binary network.
How to binarize input images for further bit-wise computation remains a difficult and open question.
Moreover, specialized hardware and deep learning frameworks that are suitable for computations of binary network are needed.

\section{Conclusions}

In this paper, we introduce a network binarization strategy for super resolution, without losing much per-pixel accuracy. 
Inspired by the structure and statistics on the gradient histogram of Laplacian pyramid, we argue that it is appropriate to pose binarization components in residual architectures, and assign a learnable parameter to each binary convolutional filter.
Qualitative and quantitative experiments have shown that residual-based SR networks with binarized components can generate comparable results to their real-weight counterparts, and obtain significant improvement in model size and computational complexity. 

{\small
\bibliographystyle{ieee}
\bibliography{egbib}
}

\end{document}